\documentclass{article}

\PassOptionsToPackage{numbers,compress}{natbib}
\usepackage[preprint]{neurips_2026}

\workshoptitle{NeurIPS 2026 Workshop on Agentic Systems for Molecular Sciences}

\usepackage{tabularx}
\usepackage[utf8]{inputenc} % allow utf-8 input
\usepackage[T1]{fontenc}    % use 8-bit T1 fonts
\usepackage{hyperref}       % hyperlinks
\usepackage{url}            % simple URL typesetting
\usepackage{booktabs}       % professional-quality tables
\usepackage{amsfonts}       % blackboard math symbols
\usepackage{nicefrac}       % compact symbols for 1/2, etc.
\usepackage{microtype}      % microtypography
\usepackage{xcolor}         % colors
\usepackage{array}
\usepackage{tabularx}
\usepackage{algorithm}
\usepackage{algorithmic}
\usepackage{multirow}
\usepackage{makecell}
\usepackage{booktabs}
\usepackage{multirow}
\usepackage{adjustbox}
\usepackage{xcolor}
\usepackage{pifont}
\usepackage{colortbl}

\usepackage{booktabs}
\usepackage{amsmath}
\usepackage{amssymb}
\usepackage{mathtools}
\usepackage{amsthm}
\usepackage{tabularx}
\theoremstyle{plain}

\theoremstyle{definition}

\theoremstyle{remark}

\newcommand{\yes}{\(\checkmark\)}
\newcommand{\no}{\(\times\)}
\usepackage[table]{xcolor}
\usepackage{enumitem}
\definecolor{stronggreen}{HTML}{A9D6B5}
\definecolor{midgreen}{HTML}{CDE8D3}
\definecolor{lightgreen}{HTML}{EAF5EC}
\definecolor{lightred}{HTML}{F4CCCC}
\newcolumntype{Y}{>{\centering\arraybackslash}X}

\title{Do Perturbation Predictors Learn Gene Interactions? Diagnosing Extrapolation in Agentic Screening}

\author{%
  Ziyuan Gao \\
  Department of Computer Science and Technology\\
  University of Cambridge\\
  \texttt{zg344@cam.ac.uk} \\
}

\begin{document}

\maketitle

\begin{abstract}
High-throughput perturbation screens increasingly use machine-learning models to predict the effects of gene combinations that were never measured. A difficult case arises when both genes in a test pair were seen individually during training, but their combination was not. In this setting, standard OOD methods often fail because neither gene appears unfamiliar on its own. We therefore compare each model prediction with a simple additive prediction obtained by summing the two observed single-gene effects. The difference between them, which we call additive disagreement (AD), measures how much extra interaction effect the model introduces for an unseen combination. Across eight prediction methods and three Perturb-seq datasets, specialised neural predictors do not consistently outperform this simple additive baseline. More importantly, predictions with larger deviations from the additive baseline tend to have larger errors. Conventional output-based and embedding-based confidence signals are inconsistent across datasets and models, whereas AD remains informative across all evaluated compositional settings. We further study how this signal can guide decisions in an agentic perturbation-screening workflow. The code is available at https://github.com/zygao930/OOD.

% These results show 
% that reliable virtual cell deployment requires 
% mechanisms for identifying unsupported interaction 
% extrapolations.
\end{abstract}

\section{Introduction}

% 1. Big picture
High-throughput perturbation screens have become a central experimental
paradigm in functional genomics, enabling systematic measurement of how
genetic interventions reshape cellular states at single-cell
resolution~\cite{dixit2016, norman2019, replogle2022, frangieh2021}.
These datasets have driven the development of virtual cell models that
aim to predict cellular responses to perturbations not directly observed
during training~\cite{roohani2023, lotfollahi2023, cui2024, hao2024}.
As such models are increasingly used to prioritise candidate
interventions and narrow experimental search
spaces~\cite{bunne2024virtual, iterpert2024},
and to support emerging
agentic workflows that autonomously design perturbation experiments,
query predictors, and rank candidates for
follow-up~\cite{roohani2024agent, perturboagent2025, boiko2023,
bran2024, huang2024crisprgpt}, their reliability
becomes a practical concern rather than merely an academic one. When
high-confidence but unreliable predictions are used to guide costly
follow-up experiments, they can, in the worst case, waste limited resources, distort
candidate rankings, and lead subsequent analyses towards misleading
biological
interpretations~\cite{vinastorne2025, miller2025, bendidi2024,
li2025benchmark}.

% 2. Current research and gap
Recent perturbation predictors include graph-based models~\cite{roohani2023}, compositional approaches~\cite{lotfollahi2023}, and single-cell foundation models~\cite{cui2024, hao2024, theodoris2023}, yet benchmarks show that these specialised methods do not consistently outperform simple linear baselines on unseen perturbations and combinations~\cite{ahlmanneltze2025, bendidi2024, wenteler2024, kernfeld2024}. Standard OOD and uncertainty methods, including maximum softmax probability~\cite{hendrycks2017}, kNN~\cite{sun2022}, Mahalanobis distance~\cite{lee2018}, deep ensembles~\cite{lakshminarayanan2017}, MC dropout~\cite{gal2016}, and selective prediction~\cite{geifman2017}, are not designed for \emph{compositional} OOD~\cite{lake2018, keysers2020}, where familiar components form an unseen combination. In perturbation prediction, both genes may therefore be familiar while their joint effect is not, making unreliable extrapolation difficult to detect from input proximity alone.

% 3. What we did
To understand where prediction errors originate, we decompose each
model's output into a component supported by observed single-gene
effects and a residual interaction term that the model extrapolates
without direct training signal. The magnitude of this residual, which
we call \emph{additive disagreement} (AD), quantifies how strongly a
prediction departs from an experimentally grounded additive reference.
We evaluate this decomposition across eight prediction methods spanning
six modelling paradigms on two combinatorial perturbation datasets,
Norman and Joung--Zhang, together with Adamson as a complementary
single-gene OOD setting~\cite{norman2019, adamson2016,
joungzhang2023}. We compare AD with eight alternative reliability
signals spanning embedding-based, output-based, and ensemble-based
approaches. We further evaluate the practical utility of these reliability signals
in an agentic perturbation-screening setting, examining whether they can
identify predictions that rely on unsupported interaction extrapolation.

Our main findings are: (1) we show across eight prediction methods that specialised neural predictors do not consistently outperform simple additive baselines on unseen gene combinations, and identify the co-expression graph as a major source of GEARS's extrapolation failures; (2) we show that generic reliability signals, including kNN distance, Mahalanobis distance, and ensemble uncertainty, are substantially weaker and less consistent than AD under compositional shift; and (3) by separating observed additive effects from model-implied interactions, we show that AD provides the most consistent indicator of prediction failure across the two compositional datasets, suggesting that current models do not reliably learn interactions that transfer to unseen combinations.

\section{Method}
\label{sec:framework}

\begin{figure*}[t]
\centering
\includegraphics[width=\textwidth]{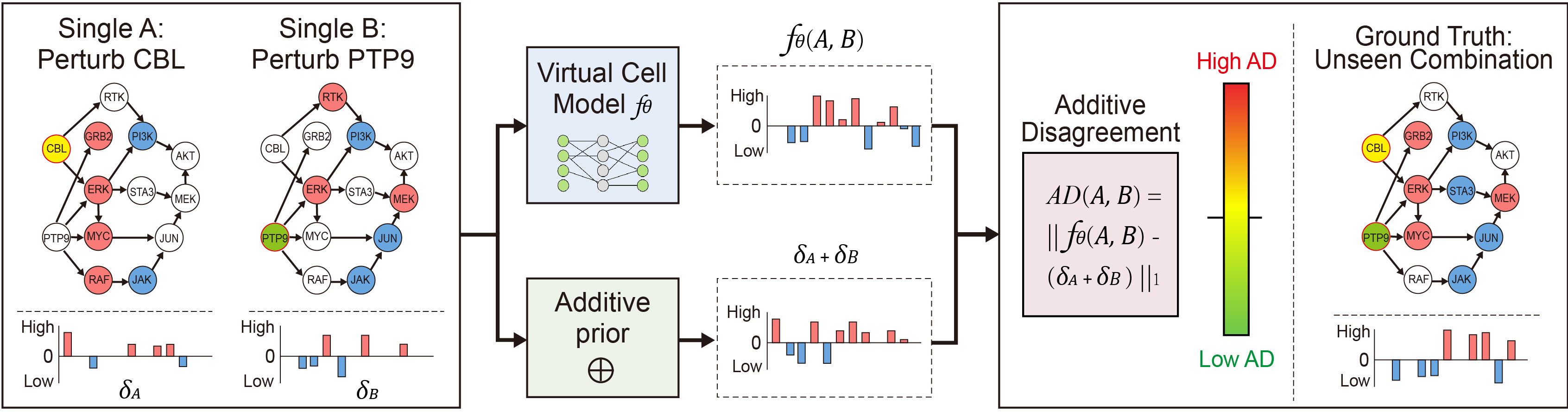}
\caption{Pipeline for diagnosing interaction extrapolation in combinatorial perturbation prediction.
}
\label{fig:pipeline}
\end{figure*}

As illustrated in Figure~\ref{fig:pipeline}, predictions for novel gene combinations contain an observed additive component and an extrapolated interaction component. Existing evaluations~\cite{ahlmanneltze2025, bendidi2024, wenteler2024, kernfeld2024} typically treat the prediction as a single point estimate, obscuring which component drives the error. We formalise this distinction to diagnose prediction failures.

% \paragraph{The compositional OOD problem.}
When a model predicts the effect of combining genes $A$ and $B$, both
genes have been observed individually during training and the model has
learned representations for each. The challenge is therefore not that
the individual inputs are unfamiliar, but that the model must predict
how $A$ and $B$ interact despite never observing this particular
combination. We evaluate this setting on CRISPRa perturbations in K562
cells from Norman et al.~\cite{norman2019} and on a second
combinatorial perturbation dataset from Joung--Zhang
et al.~\cite{joungzhang2023}. We additionally use the single-gene
CRISPRi dataset from Adamson et al.~\cite{adamson2016} to examine
whether generic reliability signals remain informative under a
different OOD setting. Dataset and split details are in Appendix~\ref{app:setup}.

% \paragraph{From problem to signal: decomposing model predictions.}
The compositional structure suggests a natural decomposition. For an
unseen pair $(A,B)$, the model prediction can be written as
\begin{equation}
  f_\theta(A,B)
  =
  \underbrace{\left(\delta_A+\delta_B\right)}
  _{\text{observed additive reference}}
  +
  \underbrace{\hat{I}_\theta(A,B)}
  _{\text{model-implied interaction}},
\end{equation}
where $\delta_A$ and $\delta_B$ denote the measured single-gene effects
and
\begin{equation}
  \hat{I}_\theta(A,B)
  =
  f_\theta(A,B)-\left(\delta_A+\delta_B\right)
\end{equation}
is the interaction correction introduced by the model.
The additive component is grounded in 
single-gene responses, whereas $\hat{I}_\theta(A,B)$ must be
extrapolated for a held-out combination. This motivates a direct
reliability signal based on the magnitude of the unsupported
correction:
\begin{equation}
  \mathrm{AD}(A,B)
  =
  \frac{1}{G}
  \left\|
  f_\theta(A,B)-\left(\delta_A+\delta_B\right)
  \right\|_1,
\end{equation}
where $G$ is the number of evaluated genes. We call this quantity
\emph{additive disagreement}. Large AD indicates that the model asserts
a strong interaction correction for a combination it has never
observed. Unlike generic uncertainty measures, AD directly targets the
component that requires compositional generalisation and has a clear
biological interpretation as the mean magnitude of the model-implied
departure from additivity.

% \paragraph{From detection to deployment.}
Standard OOD methods may fail under compositional shift because kNN~\cite{sun2022} and Mahalanobis distance~\cite{lee2018} measure input novelty, while both genes are individually observed and the novelty lies in their interaction. We therefore evaluate eight prediction methods, from Ridge and the additive prior to MLP, Transformer~\cite{vaswani2017}, CPA~\cite{lotfollahi2023}, scGPT based predictors~\cite{cui2024}, and GEARS~\cite{roohani2023}. Across models, we compare AD with eight alternative reliability signals, including output uncertainty and embedding based OOD scores. Performance is assessed by failure detection AUROC, error rank correlation, and MAE improvement under selective abstention~\cite{geifman2017} (Appendix~\ref{app:setup}).
As perturbation predictors are increasingly integrated into agentic pipelines for experimental design~\cite{roohani2024agent, perturboagent2025}, identifying predictions that rely on unsupported interaction extrapolation becomes practically important. Any system querying a predictor can simultaneously compute the additive
reference, $\delta_A + \delta_B$, and compare it with the model output
using only single-gene effect estimates.

\section{Results}
\label{sec:Results}

\subsection{Prediction Performance}

Table~\ref{tab:main-results} compares model performance on unseen gene
combinations across two compositional datasets. On Norman, Ridge
achieves the lowest full-transcriptome MAE and wins 60 of 75 evaluations
over the additive baseline. On Joung--Zhang, Ridge again achieves the
lowest MAE, but CPA wins the most pairs (29/44) and Transformer wins
20/44. Despite these differences in pairwise win rates, Additive
achieves the best Top20 MAE on Norman, while Ridge achieves the best
Top20 MAE on Joung--Zhang. GEARS and GEARS+scGPT perform worst on both
datasets and never outperform the additive baseline (0/75 on Norman,
0/44 on Joung--Zhang). Initialising with scGPT representations provides
little improvement. Overall, no neural predictor consistently
outperforms the additive baseline across both datasets and both metrics.

\begin{table*}[h]
\centering
\caption{Prediction performance on held-out gene combinations. Wins/Add. denotes the number of evaluations outperforming the additive baseline. 
}
\label{tab:main-results}
\footnotesize
\resizebox{\textwidth}{!}{%
\begin{tabular}{l ccc ccc}
\toprule
& \multicolumn{3}{c}{\textbf{Norman} (75 evaluations)}
& \multicolumn{3}{c}{\textbf{Joung--Zhang} (44 pairs)} \\
\cmidrule(lr){2-4}\cmidrule(lr){5-7}
Model & MAE $\downarrow$ & Top20 MAE $\downarrow$ & Wins/Add.
      & MAE $\downarrow$ & Top20 MAE $\downarrow$ & Wins/Add. \\
\midrule
\rowcolor{gray!20}
Additive
  & 0.0142 & \textbf{0.2121} & --
  & 0.0190 & 0.2832 & -- \\
Ridge~\cite{hoerl1970}
  & \textbf{0.0118} & 0.2312 & 60/75
  & \textbf{0.0187} & \textbf{0.2496} & 18/44 \\
MLP
  & 0.0214 & 0.3910 & 6/75
  & 0.0252 & 0.3726 & 2/44 \\
Transformer~\cite{vaswani2017}
  & 0.0215 & 0.6049 & 8/75
  & 0.0201 & 0.4051 & 20/44 \\
CPA~\cite{lotfollahi2023}
  & 0.0175 & 0.3395 & 11/75
  & 0.0192 & 0.4064 & 29/44 \\
scGPT~\cite{cui2024}
  & 0.0224 & 0.6300 & 6/75
  & 0.0209 & 0.4059 & 13/44 \\
GEARS+scGPT~\cite{roohani2023, cui2024}
  & 0.0338 & 0.2834 & 0/75
  & 0.0500 & 0.3290 & 0/44 \\
GEARS~\cite{roohani2023}
  & 0.0336 & 0.2892 & 0/75
  & 0.0512 & 0.3445 & 0/44 \\
\bottomrule
\end{tabular}%
}
\end{table*}

\subsection{Reliability Analysis}

\begin{figure*}[h]
\centering
\includegraphics[width=\textwidth]{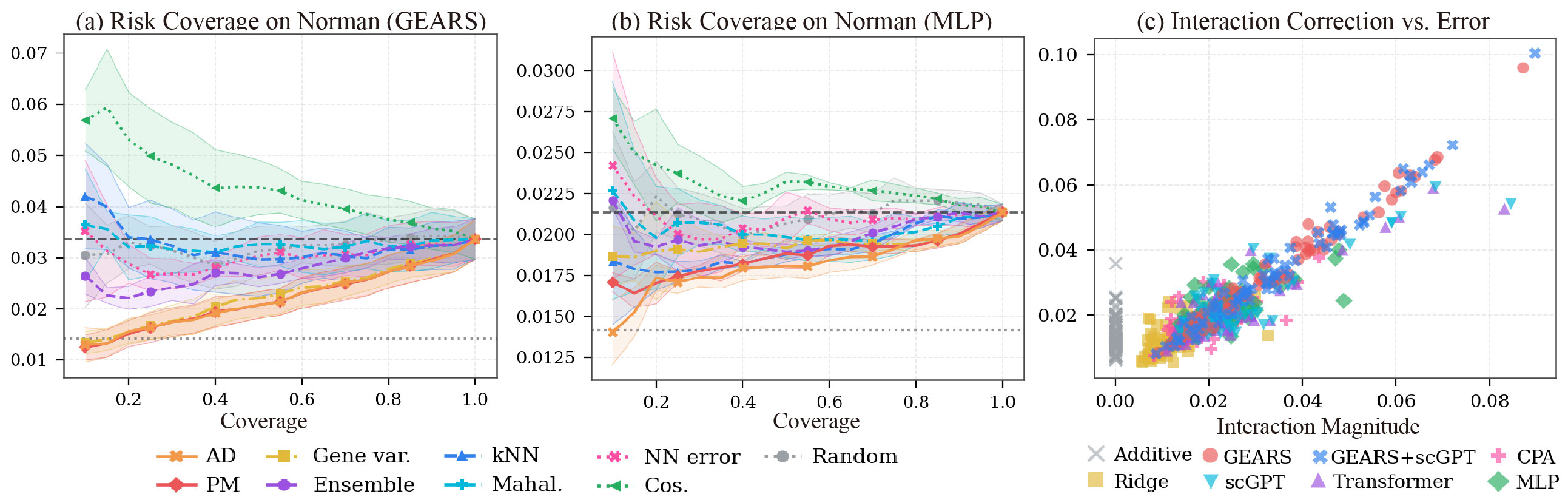}
\caption{Reliability and failure analysis on Norman. (a, b) Risk--coverage curves for GEARS and MLP across three seeds. (c) Additive disagreement correlates with true interaction magnitude.}
\label{fig:Reliability}
\end{figure*}

Table~\ref{tab:reliability-signals} compares nine reliability signals
for GEARS. Additive disagreement, prediction magnitude, and gene-wise
variance reliably identify high-error predictions, whereas kNN and ensemble-based signals are substantially weaker and Mahalanobis distance
performs near random. 
Ensemble variance is included as a model-instability baseline; its much weaker correlation with error 
than AD indicates that the failures are not explained solely by training
randomness. Because both genes are individually observed during
training, embedding distance does not directly capture the unseen
interaction under compositional OOD. Abstaining on the highest-risk
10\%, 20\%, and 50\% of predictions using additive disagreement reduces
MAE by 12\%, 19\%, and 38\%, respectively, as shown in
Figure~\ref{fig:Reliability}. AD achieves the highest AUROC at all three
failure thresholds and reaches 0.996 for the top 10\% highest-error
evaluations. Overall, compositional errors are more closely associated
with unsupported interaction corrections than with input-space novelty (full results in Appendix~\ref{app:signals}).

\begin{table}[h]
\centering
\caption{Reliability signals across failure thresholds and abstention rates.
Top-$k\%$ denotes the highest-error predictions.}
\label{tab:reliability-signals}
\small
\renewcommand{\arraystretch}{1}
\begin{tabularx}{\linewidth}{@{}l*{7}{Y}@{}}
\toprule
& \multicolumn{3}{c}{Failure AUROC $\uparrow$}
& \multicolumn{1}{c}{Error ranking $\uparrow$}
& \multicolumn{3}{c}{Abstention improvement $\uparrow$} \\
\cmidrule(lr){2-4}\cmidrule(lr){5-5}\cmidrule(lr){6-8}
Signal & Top50\% & Top25\% & Top10\% & Spearman & 10\% & 20\% & 50\% \\
\midrule
\rowcolor{gray!20}
Ensemble variance
& 0.670 & 0.708 & 0.782 & 0.348 & $+1\%$ & $+4\%$ & $+17\%$ \\

Additive disagreement
& \textbf{0.994} & \textbf{0.989} & \textbf{0.996}
& \textbf{0.982} & $+12\%$ & $+19\%$ & $+38\%$ \\
Prediction magnitude
& 0.993 & 0.986 & 0.983 & 0.978
& $+11\%$ & $+19\%$ & $+38\%$ \\
Gene-wise variance
& 0.933 & 0.945 & 0.983 & 0.903
& $+11\%$ & $+19\%$ & $+35\%$ \\

NN error proxy
& 0.683 & 0.710 & 0.664 & 0.345
& $+2\%$ & $+5\%$ & $+14\%$ \\
kNN distance
& 0.623 & 0.638 & 0.599 & 0.290
& $+4\%$ & $+6\%$ & $+16\%$ \\
Mahalanobis distance
& 0.491 & 0.463 & 0.418 & $-0.081$
& $+2\%$ & $-2\%$ & $-7\%$ \\
Cosine to nearest
& 0.175 & 0.066 & 0.125 & $-0.752$
& $-7\%$ & $-12\%$ & $-29\%$ \\
Random
& 0.563 & 0.508 & 0.593 & 0.073
& $-2\%$ & $+1\%$ & $+6\%$ \\
\bottomrule
\end{tabularx}
\end{table}

%%%%%%%%%%%%%%%%%%%%%%%%%%%%%%%   Failure Analysis    %%%%%%%%%%%%%%%%%%%%%%%%%%
\subsection{Failure Analysis}

\paragraph{Failure modes of interaction extrapolation.}

Table~\ref{tab:failure-modes} shows representative failure patterns in
GEARS predictions. We observe two recurring behaviours. In the first,
GEARS produces large interaction corrections that move predictions far
from the additive reference. In the second, the corrections are smaller
but still introduce unnecessary deviations. In both cases, the
model-implied interaction correction is closely associated with the
increase in prediction error. For example, on CEBPE+CEBPA the additive
baseline achieves an MAE of 0.036, whereas GEARS produces an

\begin{table*}[h]
\centering
\caption{Representative failure cases. Left: high-interaction pairs
where GEARS produces large unsupported interaction corrections.
Right: lower-interaction pairs where GEARS adds unnecessary noise.
AD = additive disagreement.}
\label{tab:failure-modes}
\renewcommand{\arraystretch}{1}
\footnotesize
\setlength{\tabcolsep}{4pt}
\begin{tabular}{p{2.8cm}p{1.1cm}p{1.0cm}p{1.0cm}|p{2.8cm}p{1.1cm}p{1.0cm}p{1.0cm}}
\toprule
\multicolumn{4}{c|}{\textbf{High-interaction failures}}
& \multicolumn{4}{c}{\textbf{Lower-interaction noise}} \\
\midrule
Pair & GEARS & Add. & \textbf{AD}
& Pair & GEARS & Add. & \textbf{AD} \\
\midrule
CEBPE+CEBPA & \textbf{0.096} & 0.036 & \textbf{0.087}
& SET+CEBPE & \textbf{0.042} & 0.025 & \textbf{0.038} \\
JUN+CEBPA & 0.068 & 0.026 & 0.069
& PRDM1+CBFA2T3 & 0.041 & 0.020 & 0.038 \\
CEBPE+CEBPB & 0.068 & 0.023 & 0.068
& SGK1+S1PR2 & 0.036 & 0.013 & 0.039 \\
C3orf72+FOXL2 & 0.064 & 0.022 & 0.061
& CEBPE+CNN1 & 0.036 & 0.018 & 0.032 \\
ZBTB10+SNAI1 & 0.063 & 0.021 & 0.063
& ETS2+IKZF3 & 0.032 & 0.013 & 0.035 \\
\bottomrule
\end{tabular}
\end{table*}

MAE of
0.096 with an AD of 0.087, nearly tripling the error. Conversely, on
SET+CEBPE the additive baseline already performs better at 0.025, yet
GEARS introduces a smaller correction with AD $=0.038$. See Appendix~\ref{app:mechanism} for correction magnitudes and GEARS ablations.

\begin{figure*}[h]
\centering
\includegraphics[width=\textwidth]{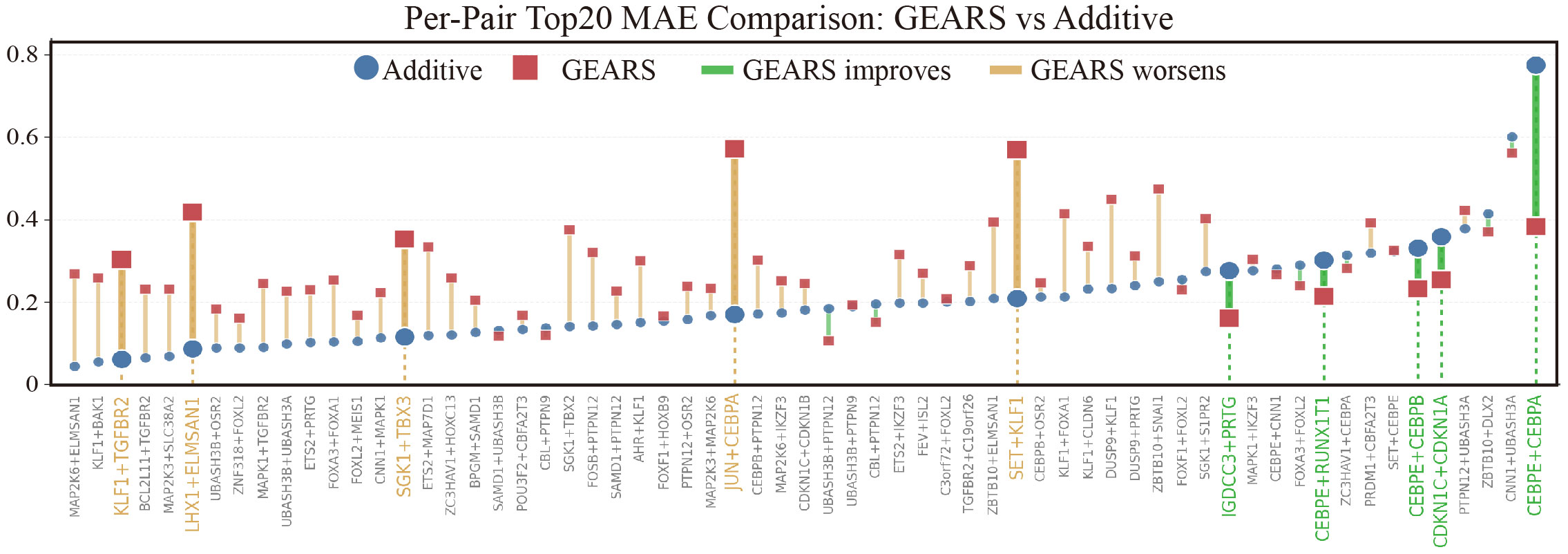}
\caption{Per-pair Top20 MAE for GEARS and the additive baseline across 75 Norman evaluations. Blue circles and red squares denote additive and GEARS errors, respectively; connectors show their difference, with green indicating GEARS improvement and orange indicating worse performance.}
\label{fig:gene_level}
\end{figure*}

\paragraph{Gene-level analysis.}
We decompose pair-level predictions into per-gene contributions to identify where GEARS's corrections fail. Larger corrections are strongly associated with higher error, suggesting that they often add noise rather than transferable interaction signal. GEARS assigns $9.7\times$ larger corrections to the top-20 DE genes than to the rest of the transcriptome (mean $|\hat{I}_g|=0.324$ vs.\ $0.033$), even though the additive baseline already captures much of their signal. Because these graph-based corrections do not distinguish observed from unseen combinations, they often shift predictions away from an informative additive estimate. Accordingly, GEARS improves over the additive baseline on only 25.3\% of evaluated pairs under Top20 MAE (Figure~\ref{fig:gene_level}).

\section{Discussion}

We test whether the observed failures arise from weak interaction signal, interaction complexity, model design, or evaluation choice. The failures are not due to weak interactions: the interaction component accounts for 68\% of the total effect on Norman and 104\% on average on Joung--Zhang, with no pair below 10\%. Interaction magnitude also does not consistently predict difficulty. GEARS excess error increases with interaction magnitude on Norman ($\rho = 0.527$, $p < 10^{-4}$), whereas several other models show weak or negative correlations, and all evaluated models show strong negative correlations on Joung--Zhang, including GEARS ($\rho = -0.749$, $p < 10^{-4}$). The interactions nevertheless exhibit shared structure: on Norman, the first principal component explains 31.3\% of the variance, four explain 50\%, and 21 explain 90\%; on Joung--Zhang, the corresponding values are 21.3\%, four, and 15 components. Thus, interactions are substantial and structured, but their difficulty depends on both model and dataset rather than interaction strength alone.

The Norman analyses further implicate the co-expression graph in GEARS's failures. Co-expression degree correlates with both predicted interaction magnitude ($\rho = 0.633$) and prediction error ($\rho = 0.626$). Removing the co-expression graph reduces MAE from 0.0336 to 0.0228, while removing both graphs yields 0.0259. This suggests that co-expression message passing amplifies erroneous corrections around highly connected genes rather than capturing transferable pair-specific interactions, although graph-ablated variants still underperform the additive baseline.

Overall, the evidence points primarily to modeling choices. The evaluated architectures do not consistently generalize structured interaction effects to unseen combinations, and in GEARS, co-expression message passing is a major contributor to failure. We also find an evaluation-dependent limitation of additive disagreement: on Norman, its failure-detection AUROC falls from 0.990 under transcriptome-wide MAE to 0.619 under Top20 MAE, showing that reliability over the full transcriptome does not necessarily transfer to a restricted gene set.

\clearpage
{\small
\bibliographystyle{unsrtnat}
\bibliography{refs}
}

\clearpage
\appendix
\appendix
\section{Additional Analysis of Reliability Signals}
\label{app:theory}

\paragraph{Why embedding-based novelty can miss compositional shift.}
Under compositional shift, both constituent genes $A$ and $B$ have
been observed individually during training, although their combination
has not. Pair representations constructed from familiar gene embeddings
may therefore remain close to the training distribution. This is
particularly likely when pair embeddings are formed through smooth
aggregation functions such as averaging or pooling, because the
resulting representation primarily reflects component-level familiarity
rather than whether the interaction has been observed. Consequently,
kNN distance, Mahalanobis distance, and cosine similarity may assign
low novelty to an unseen combination even when its interaction
prediction is unreliable. Embedding-based detection need not fail in
every setting, but component-level familiarity weakens the alignment
between representation-space novelty and interaction-level novelty.

\paragraph{Relationship between AD and prediction error.}
Let the true interaction residual be
\begin{equation}
I_{\mathrm{true}}(A,B)
=
y(A,B)-(\delta_A+\delta_B).
\end{equation}
Since
\begin{equation}
f_\theta(A,B)-y(A,B)
=
\hat{I}_\theta(A,B)-I_{\mathrm{true}}(A,B),
\end{equation}
the triangle inequality gives
\begin{equation}
\left|
\|\hat{I}_\theta(A,B)\|_{1,G}
-
\|I_{\mathrm{true}}(A,B)\|_{1,G}
\right|
\leq
\|f_\theta(A,B)-y(A,B)\|_{1,G}
\leq
\|\hat{I}_\theta(A,B)\|_{1,G}
+
\|I_{\mathrm{true}}(A,B)\|_{1,G}.
\end{equation}
In the additive limiting case,
$I_{\mathrm{true}}(A,B)=0$, prediction error is exactly equal to AD.
More generally, AD measures how strongly the model departs from the
experimentally grounded additive reference. A large AD does not
necessarily imply a large error, because a large correction may be
accurate if it matches the true interaction. AD should therefore be
interpreted as a risk signal for unsupported interaction extrapolation,
rather than as a deterministic estimate of prediction error.

Empirically, AD is strongly associated with GEARS prediction error on
both compositional datasets, with Spearman correlations of
$\rho=0.982$ on Norman and $\rho=0.830$ on Joung--Zhang. Across all
evaluated models, its correlations range from $0.528$ to $0.982$ on
Norman and from $0.767$ to $0.939$ on Joung--Zhang. Thus, although the
bounds do not guarantee a monotonic relationship, large model-implied
interaction corrections are consistently associated with higher error
in the evaluated compositional settings.

\paragraph{Prediction magnitude versus additive disagreement.}
Because
\begin{equation}
f_\theta(A,B)
=
(\delta_A+\delta_B)+\hat{I}_\theta(A,B),
\end{equation}
PM and AD may induce similar rankings when variation in the additive
component is small relative to variation in the model-implied
interaction correction. This occurs for GEARS on Norman, where their
Spearman correlations with error are $0.978$ and $0.982$,
respectively.
However, PM also increases for perturbations with large but
experimentally supported single-gene effects. It therefore conflates
the magnitude of the additive response with the magnitude of the
extrapolated interaction correction. This distinction becomes clearer
across models and datasets. On Joung--Zhang, PM correlates strongly
with error for Ridge ($\rho=0.812$), but is close to zero for
Transformer ($\rho=0.041$) and scGPT ($\rho=0.021$). AD remains
positively associated with error for every evaluated model. By
subtracting the observed additive component, AD more specifically
targets interaction-level extrapolation.

\paragraph{Dependence on the evaluation metric.}
The reliability of AD depends on how prediction error is defined. On
Norman, AD achieves a failure-detection AUROC of $0.990$ when failure
is measured by full-transcriptome MAE, but only $0.619$ under Top20
MAE. The same pattern holds for squared error, with AUROC decreasing
from $0.961$ under full-transcriptome MSE to $0.597$ under Top20 MSE.
This difference follows from the gene sets used by the signal and the
evaluation metric. Full-transcriptome AD aggregates departures from
the additive reference across all evaluated genes and is therefore
well aligned with full-transcriptome error. Top20 metrics evaluate only
the most differentially expressed genes for each pair. A prediction
may accumulate small deviations across many genes while remaining
accurate on the Top20 set, or may have a modest global deviation but
large errors on a few selected genes. Reliability established at the
transcriptome-wide level therefore does not automatically transfer to
a restricted gene set. Applications focused on biomarkers, pathways,
or selected genes should define and validate reliability over the same
target set.

%%%%%%%%%%%%%%%%%%%%%%%%%%%%%%%%%%%%%%%%%%%%%%%%%%%%%%%%%%%%%%%%%%%%
\newpage
\section{Experimental Details}
\label{app:setup}

\subsection{Datasets}

\paragraph{Datasets.}
We evaluate on three Perturb-seq datasets.

\textbf{Norman (CRISPRa, K562).}
After filtering through the GEARS preprocessing pipeline, the dataset contains 89,357 cells and 5,045 genes~\cite{roohani2023}. It includes 105 single-gene and 131 combinatorial perturbation conditions~\cite{norman2019}. Among the combinatorial conditions, 128 pairs have both constituent genes represented among the observed single-gene perturbations and are therefore eligible for compositional OOD evaluation. This provides the primary setting in which both constituent genes are observed during training but their combination is held out.

\textbf{Joung--Zhang (combinatorial).}
We use the combinatorial perturbation dataset from Joung et al.~\cite{joungzhang2023}. After requiring both constituent genes of a double perturbation to have corresponding single-gene conditions, 44 pairs remain eligible for evaluation. This provides a second compositional OOD setting with a different perturbation library and experimental context.

\textbf{Adamson (CRISPRi, K562).}
We use the single-gene perturbation dataset from Adamson et al.~\cite{adamson2016} as a complementary single-gene OOD setting. Here, the held-out perturbation gene itself is absent from training. The evaluation contains 48 held-out single-gene perturbations across three splits.

All three datasets (Norman, Joung--Zhang, Adamson) are publicly available
and cited to their original publications, and the model implementations
we build on (GEARS, scGPT, CPA) are used under their respective
open-source licenses.

\subsection{Preprocessing}

For each perturbation condition $c$, we represent its effect relative to control as
\begin{equation}
\delta_c = \mu_c-\mu_{\mathrm{ctrl}},
\end{equation}
where $\mu_c$ and $\mu_{\mathrm{ctrl}}$ denote the mean expression profiles of the perturbation and control cells, respectively. A double perturbation $(A,B)$ is eligible for compositional evaluation only when both corresponding single-gene conditions are observed, allowing the additive reference
\begin{equation}
\delta_A+\delta_B
\end{equation}
to be constructed.

Ridge, MLP, Transformer, CPA, and scGPT are trained on condition-level mean effect vectors, whereas GEARS is trained on single-cell expression profiles through its original data pipeline. For Top20 MAE, the 20 genes with the largest absolute observed effect are selected separately for each held-out perturbation.

\paragraph{Splits.}
For Norman, each seed $(42,43,44)$ partitions the eligible pairs into 20\% test pairs, 20\% validation pairs, and 60\% training pairs. Each split contains 25 test pairs and 25 validation pairs, yielding 75 pair-seed test evaluations across the three seeds. All single-gene conditions and non-eligible combinatorial conditions remain in training.

For Joung--Zhang, each of the 44 eligible double perturbations is held out once in leave-one-out cross-validation. All remaining conditions are used for training, and each model is retrained independently for every fold.

For Adamson, single-gene perturbations are held out across three predefined splits, yielding 48 test evaluations in total.

\newpage
\subsection{Prediction Methods}
\label{app:models}

All experiments were run on an internal cluster of 10 NVIDIA A100 GPUs.

\begin{table}[h]
\centering
\caption{Prediction method details. Non-graph models are trained on condition-level mean perturbation effects. GEARS is trained on single-cell expression profiles using its original data pipeline.}
\label{tab:model-details}
\small
\renewcommand{\arraystretch}{1.2}
\begin{tabularx}{\linewidth}{@{}>{\raggedright\arraybackslash}p{2.0cm}>{\raggedright\arraybackslash}p{3.0cm}>{\raggedright\arraybackslash}X@{}}
\toprule
Model & Paradigm & Key details \\
\midrule
Additive & Prior & $\hat{\delta}_{A+B}=\delta_A+\delta_B$; no training \\
Ridge~\cite{hoerl1970} & Linear & Multi-output regression with one-hot perturbation indicators and $L_2$ regularization, $\alpha=1.0$ \\
MLP & Feedforward & Learned 64-dimensional gene embeddings, two 256-dimensional hidden layers, ReLU, LayerNorm, dropout 0.1, learning rate $10^{-3}$, and a maximum of 200 epochs \\
Transformer~\cite{vaswani2017} & Attention & 64-dimensional embeddings, 4 attention heads, 2 encoder layers, 256-dimensional feedforward blocks, pre-normalization, learnable positional embeddings, and mean pooling \\
CPA~\cite{lotfollahi2023,kingma2014} & Compositional VAE & 64-dimensional latent representation with additive perturbation embeddings, KL-weight annealing to 0.1 over 100 epochs, a maximum of 300 epochs, and patience 40 \\
scGPT~\cite{cui2024} & Foundation & Frozen pretrained 512-dimensional scGPT gene embeddings projected to 128 dimensions, followed by a trainable 3-layer, 4-head Transformer predictor with AdamW learning rate $5{\times}10^{-4}$ \\
GEARS+scGPT & Foundation+Graph & GEARS with gene embeddings initialized from scGPT representations reduced from 512 to 64 dimensions using principal component analysis \\
GEARS~\cite{roohani2023} & Graph+biology & Gene Ontology similarity and co-expression graphs, hidden dimension 64, 15 epochs, learning rate $10^{-3}$, and batch size 64 \\
\bottomrule
\end{tabularx}
\end{table}

\subsection{Reliability Signals}
\label{app:signals}

All magnitude-based signals are averaged over the $G$ evaluated genes so that their scale is comparable to MAE.

\begin{table}[h]
\centering
\caption{Reliability signal definitions. Each signal produces one scalar risk score per prediction, with higher values indicating greater predicted risk.}
\label{tab:signal-details}
\small
\setlength{\tabcolsep}{4pt}
\renewcommand{\arraystretch}{1.2}
\begin{tabularx}{\linewidth}{@{}>{\raggedright\arraybackslash}p{3.2cm}>{\raggedright\arraybackslash}p{1.8cm}>{\raggedright\arraybackslash}X@{}}
\toprule
Signal & Source & Definition \\
\midrule
Additive disagreement & Prior & $\displaystyle \mathrm{AD}(A,B)=\frac{1}{G}\left\|f_\theta(A,B)-(\delta_A+\delta_B)\right\|_1$ \\
Prediction magnitude & Output & $\displaystyle \mathrm{PM}(A,B)=\frac{1}{G}\left\|f_\theta(A,B)\right\|_1$ \\
Gene-wise variance & Output & $\displaystyle \mathrm{Var}_{g}\left[f_\theta(A,B)_g\right]$ \\
Ensemble variance~\cite{lakshminarayanan2017} & Output & $\displaystyle \frac{1}{G}\sum_{g=1}^{G}\mathrm{Var}_{s}\left[f_{\theta_s}(A,B)_g\right]$ across three independently trained models \\
kNN distance~\cite{sun2022} & Embedding & Mean cosine distance to the five nearest training perturbations in the corresponding model representation space \\
Mahalanobis distance~\cite{lee2018} & Embedding & Mahalanobis distance from the empirical distribution of training representations \\
NN error proxy & Embedding & Prediction error associated with the nearest training perturbation in representation space \\
Cosine to nearest & Embedding & Negated cosine similarity to the nearest training perturbation \\
Random & Baseline & Independent score sampled uniformly from $[0,1]$ \\
\bottomrule
\end{tabularx}
\end{table}

%%%%%%%%%%%%%%%%%%%%%%%%%%%%%%%%%%%%%%%%%%%%%%%%%%%%%%%%%%%%%%%%%%%%%%%%%%%%%%
\clearpage
\section{Additional Analysis of Selective Prediction}
\label{app:selective}

This section provides complete selective-prediction results and
additional analyses of the model-implied interaction corrections.

\subsection{Why Models Extrapolate Unreliably}
\label{app:mechanism}

\subsubsection{Magnitude of Model-Implied Interaction Corrections}

For each model, we measure the mean absolute interaction correction
introduced relative to the additive reference.
Table~\ref{tab:interaction-corrections} reports the distribution of
these corrections on Norman and their Spearman correlation with
prediction error.
Ridge introduces the smallest corrections and achieves the lowest
overall MAE. The neural predictors generally introduce larger
corrections. GEARS and GEARS+scGPT produce the largest corrections,
with mean AD values of 0.0344 and 0.0348, respectively. Their
corrections are also strongly ranked with prediction error
($\rho=0.982$ and $\rho=0.981$). This indicates that the main GEARS
failures occur when the model moves strongly away from the additive
reference without transferable pair-specific evidence.

\begin{table}[h]
\centering
\caption{Model-implied interaction corrections on Norman. Mean, median,
and 90th percentile report the distribution of additive disagreement
across 75 test evaluations. Correlation is the Spearman correlation
between AD and prediction error.}
\label{tab:interaction-corrections}
\small
\setlength{\tabcolsep}{5pt}
\renewcommand{\arraystretch}{1.1}
\begin{tabular}{lcccc}
\toprule
Model & Mean AD & Median AD & 90th percentile & Corr.\ with error \\
\midrule
Additive & 0 & 0 & 0 & -- \\
Ridge & 0.0113 & 0.0096 & 0.0155 & 0.528 \\
CPA & 0.0184 & 0.0170 & 0.0249 & 0.585 \\
MLP & 0.0227 & 0.0208 & 0.0297 & 0.660 \\
Transformer & 0.0241 & 0.0212 & 0.0344 & 0.627 \\
scGPT & 0.0251 & 0.0218 & 0.0356 & 0.598 \\
GEARS+scGPT & 0.0348 & 0.0324 & 0.0589 & 0.981 \\
GEARS & 0.0344 & 0.0314 & 0.0597 & \textbf{0.982} \\
\bottomrule
\end{tabular}
\end{table}

\paragraph{GEARS ablations.}
We ablate the graph components and training composition of GEARS to identify
the source of unreliable interaction extrapolation 
(Table~\ref{tab:gears-ablation}). Removing the co-expression graph gives the
largest improvement and reduces MAE from 0.0336 to 0.0228. By contrast,
removing the GO graph slightly worsens performance.
In GEARS, each gene aggregates information from its neighbours during message
passing. Highly connected genes therefore receive signals from many sources.
Interaction correction magnitude correlates with node degree
($\rho = 0.633$), showing that highly connected genes tend to receive larger
corrections. These aggregated signals can fit interaction patterns in the
training pairs. However, the graph is fixed and does not record which gene
combinations have been observed. GEARS therefore applies the same mechanism to
held-out pairs, even when no pair-specific evidence is available. This process
transfers interaction patterns from the training set to unseen combinations
and often increases prediction error.

\begin{table}[h]
\centering
\caption{GEARS ablations. \yes/\no indicate whether each component is used.
Pairs denotes combinatorial training data; CoExp denotes co-expression.}
\label{tab:gears-ablation}
\small
\setlength{\tabcolsep}{4pt}
\renewcommand{\arraystretch}{1}
\begin{tabular}{lcccccc}
\toprule
Configuration & Pairs & GO & CoExp & MAE $\downarrow$ & Wins/Add. \\
\midrule
Full GEARS         & \yes & \yes & \yes & 0.0336 & 0/75 \\
No GO              & \yes & \no  & \yes & 0.0342 & 0/75 \\
No CoExp           & \yes & \yes & \no  & 0.0228 & 3/75 \\
No graphs          & \yes & \no  & \no  & 0.0259 & 3/75 \\
Singles-only GEARS & \no  & \yes & \yes & 0.0291 & 1/75 \\
\addlinespace[2pt]
\rowcolor{gray!20}
Additive           & --   & --   & --   & \textbf{0.0142} & -- \\
\bottomrule
\end{tabular}
\end{table}

Training on single-gene perturbations alone already produces a higher MAE than
the additive baseline (0.0291 vs.\ 0.0142). The model therefore does not recover
additive predictions when interaction supervision is absent. Adding
combinatorial training data further increases MAE to 0.0336. This result
suggests that GEARS learns interaction corrections from the training pairs but
does not transfer them reliably to unseen pairs. Overall, the main source of
error is not simply a failure to model additive effects, but the introduction
of interaction corrections without transferable evidence.

\subsection{Reliability Signals for Selective Prediction}
\label{app:signals}

Having established that unreliable predictions arise from interaction
corrections without transferable evidence, we now evaluate whether
reliability signals can rank these failures. We report fixed-abstention
results on Norman, full risk--coverage curves across OOD settings, and
the transfer of each signal across datasets.

\subsubsection{Selective Prediction on Norman}

For each reliability signal, we rank the 75 GEARS test evaluations by
predicted risk, remove the highest-risk fraction, and recompute MAE on
the retained evaluations. Table~\ref{tab:abstention} reports results
from 10\% to 50\% abstention.
Additive disagreement, prediction magnitude, and gene-wise variance
consistently reduce retained-set error. AD reduces MAE by 19\% at
20\% abstention and by 38\% at 50\% abstention. Ensemble variance,
the nearest-neighbour error proxy, and kNN distance provide smaller
improvements. Mahalanobis distance and the cosine-to-nearest signal
can worsen performance, indicating that representation-space proximity
does not reliably identify failed interaction extrapolation.

\begin{table*}[h]
\centering
\caption{Selective prediction for GEARS on Norman. The MAE without
abstention is 0.0336. At each abstention level, the highest-risk
evaluations are removed and MAE is recomputed on the retained set.
Parentheses report the relative MAE reduction from the full-coverage
result.}
\label{tab:abstention}
\small
\setlength{\tabcolsep}{4pt}
\renewcommand{\arraystretch}{1.1}

\newcommand{\gain}[1]{\textcolor{green!50!black}{#1}}
\newcommand{\loss}[1]{\textcolor{red!70!black}{#1}}

\begin{tabular}{lccccc}
\toprule
Signal & Abs10\% & Abs20\% & Abs30\% & Abs40\% & Abs50\% \\
\midrule
Additive disagreement & 0.0296 \gain{(+12\%)} & 0.0272 \gain{(+19\%)} & 0.0249 \gain{(+26\%)} & 0.0226 \gain{(+33\%)} & 0.0207 \gain{(+38\%)} \\
Prediction magnitude & 0.0298 \gain{(+11\%)} & 0.0272 \gain{(+19\%)} & 0.0248 \gain{(+26\%)} & 0.0226 \gain{(+33\%)} & 0.0208 \gain{(+38\%)} \\
Gene-wise variance & 0.0298 \gain{(+11\%)} & 0.0273 \gain{(+19\%)} & 0.0256 \gain{(+24\%)} & 0.0246 \gain{(+27\%)} & 0.0217 \gain{(+35\%)} \\
Ensemble variance & 0.0332 \gain{(+1\%)} & 0.0321 \gain{(+4\%)} & 0.0298 \gain{(+11\%)} & 0.0292 \gain{(+13\%)} & 0.0277 \gain{(+17\%)} \\
NN error proxy & 0.0331 \gain{(+2\%)} & 0.0320 \gain{(+5\%)} & 0.0308 \gain{(+8\%)} & 0.0290 \gain{(+14\%)} & 0.0290 \gain{(+14\%)} \\
kNN distance & 0.0323 \gain{(+4\%)} & 0.0317 \gain{(+6\%)} & 0.0315 \gain{(+6\%)} & 0.0299 \gain{(+11\%)} & 0.0282 \gain{(+16\%)} \\
Mahalanobis distance & 0.0329 \gain{(+2\%)} & 0.0342 \loss{(-2\%)} & 0.0351 \loss{(-4\%)} & 0.0354 \loss{(-5\%)} & 0.0360 \loss{(-7\%)} \\
Cosine to nearest & 0.0359 \loss{(-7\%)} & 0.0376 \loss{(-12\%)} & 0.0398 \loss{(-18\%)} & 0.0405 \loss{(-21\%)} & 0.0432 \loss{(-29\%)} \\
Random & 0.0344 \loss{(-2\%)} & 0.0334 \gain{(+1\%)} & 0.0315 \gain{(+6\%)} & 0.0314 \gain{(+7\%)} & 0.0316 \gain{(+6\%)} \\
\bottomrule
\end{tabular}
\end{table*}

\newpage
\subsubsection{Risk--Coverage Across OOD Settings}

Risk--coverage curves complement the fixed abstention rates by showing
retained-set error across the full coverage range. Lower MAE at a given
coverage indicates that a signal more effectively ranks unreliable
predictions.

\begin{figure}[h]
\centering
\includegraphics[width=\linewidth]{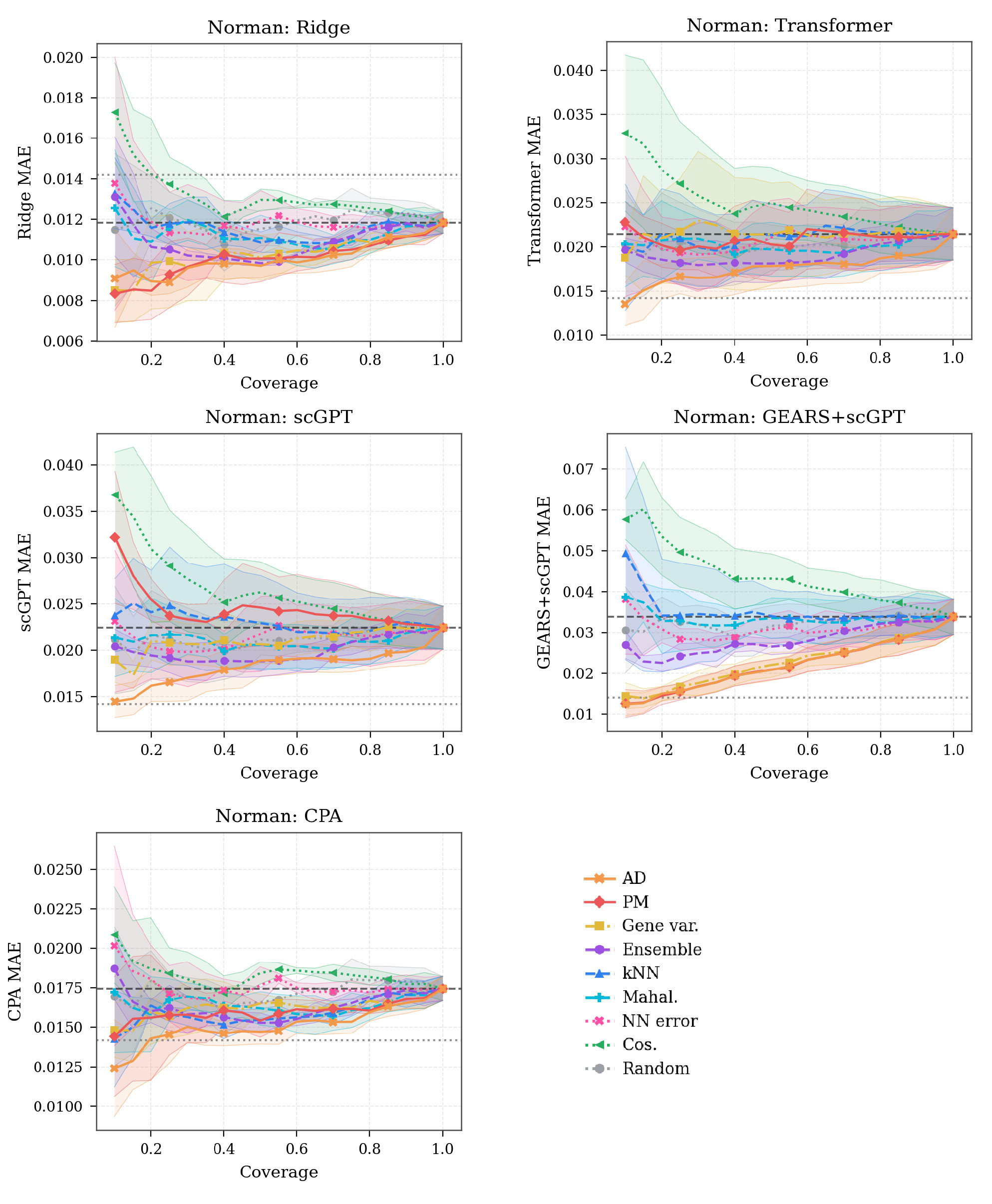}
\caption{Risk--coverage curves on Norman under compositional OOD. Each curve reports retained-set MAE as progressively higher-risk predictions are removed. Lower MAE at a given coverage indicates better selective prediction.}
\label{fig:risk-coverage-norman}
\end{figure}

\newpage
\begin{figure}[h]
\centering
\includegraphics[width=0.95\linewidth]{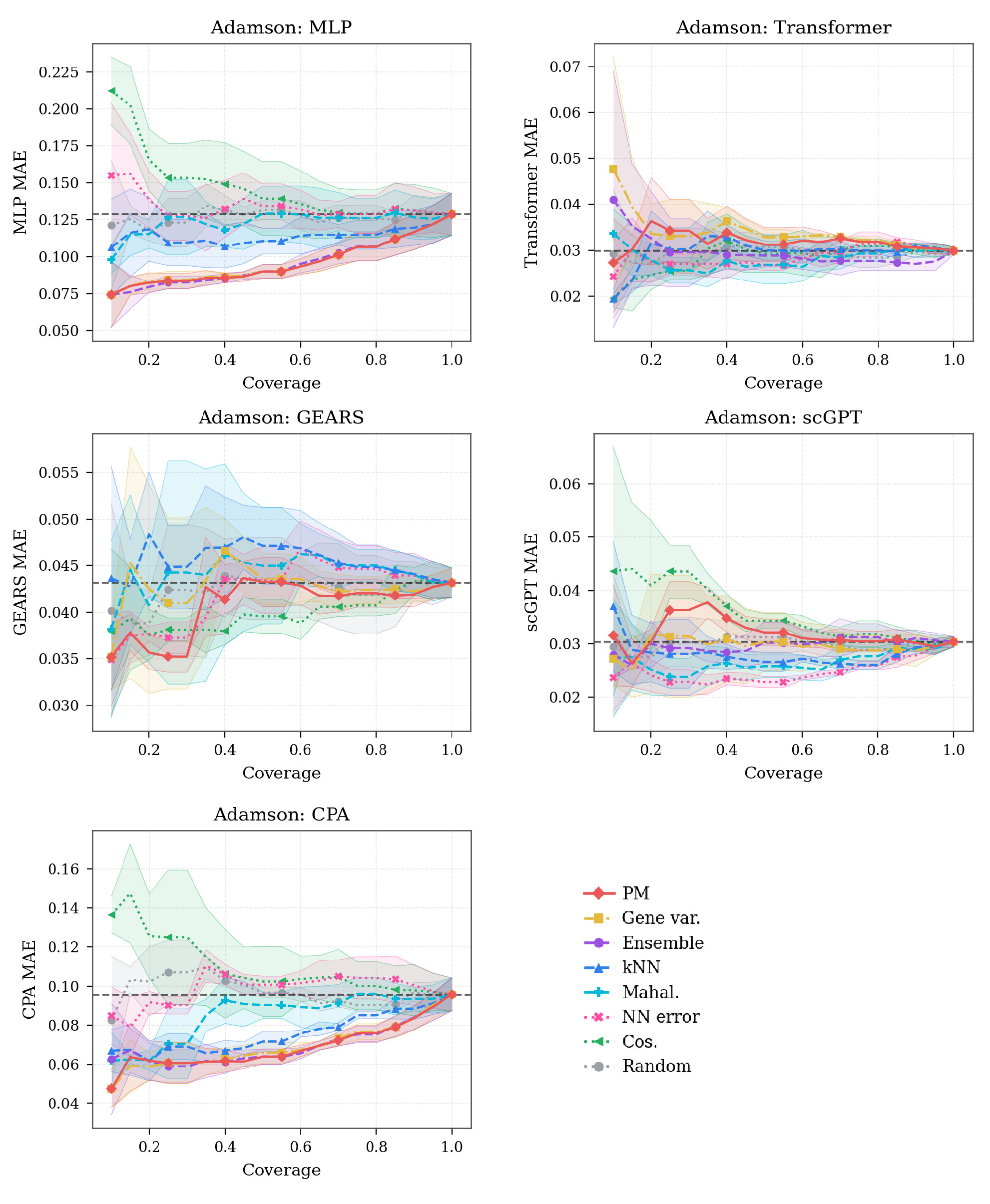}
\caption{Risk--coverage curves on Adamson under single-gene OOD. Each curve reports retained-set MAE as progressively higher-risk predictions are removed. AD is undefined because the held-out perturbation gene is unobserved.}
\label{fig:risk-coverage-adamson}
\end{figure}

\newpage
\begin{figure}[h]
\centering
\includegraphics[width=0.95\linewidth]{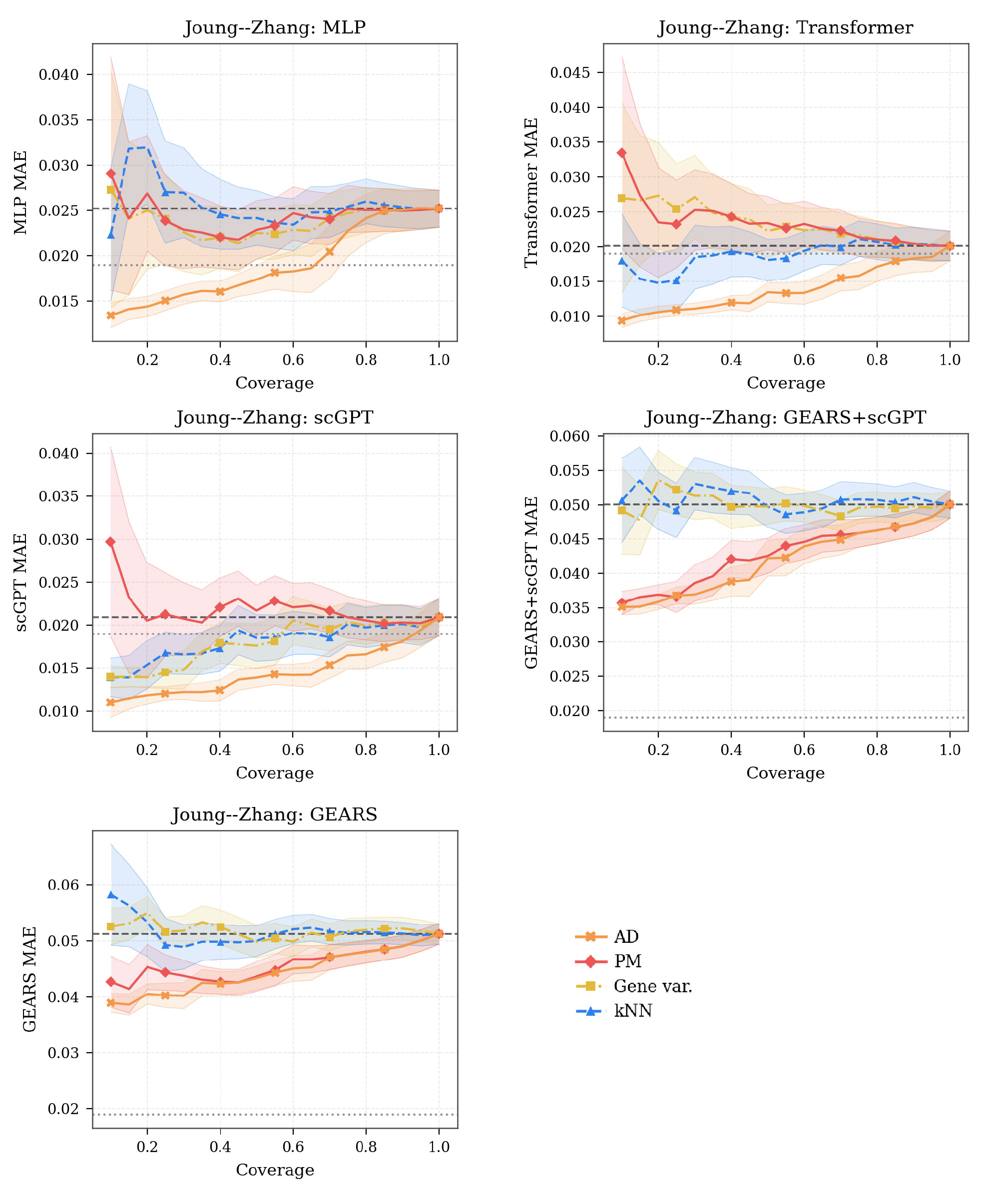}
\caption{Risk--coverage curves on Joung--Zhang under compositional OOD. Each curve reports retained-set MAE as progressively higher-risk predictions are removed. Lower MAE at a given coverage indicates better selective prediction.}
\label{fig:risk-coverage-joungzhang}
\end{figure}

\clearpage
\subsubsection{Cross-dataset Reliability}

Table~\ref{tab:cross-dataset} evaluates whether 
reliability signals transfer across datasets. On Norman, AD ranks errors well across all models, with the strongest correlations observed for GEARS and
GEARS+scGPT. On Joung--Zhang, AD remains consistently positive across
models, while PM and GV become substantially more variable: PM
correlates strongly with error for Ridge ($\rho = 0.812$) but is close
to zero for Transformer ($\rho = 0.041$) and scGPT ($\rho = 0.021$).
kNN distance is also inconsistent and can correlate negatively with
error. On Adamson, where the OOD shift is at the single-gene level
rather than the combination level, PM and GV are strong for MLP and
CPA, moderate for GEARS, and weak for Transformer and scGPT. In
contrast, AD maintains positive correlations with error across all
models and both compositional datasets ($\rho = 0.528$--$0.939$),
making it the most consistent ranking signal across settings.

\begin{table*}[h]
\centering
\caption{Cross-dataset reliability of uncertainty signals. Abs20 denotes improvement after 20\% abstention. AD is undefined for single-gene OOD on Adamson.}
\label{tab:cross-dataset}
\small
\renewcommand{\arraystretch}{0.95}
\resizebox{\textwidth}{!}{%
\begin{tabular}{lllrrrrc}
\toprule
 & & & \multicolumn{4}{c}{Spearman $\rho$ (signal vs.\ error) $\uparrow$}
 & \multicolumn{1}{c}{Abstention $\uparrow$} \\
\cmidrule(lr){4-7} \cmidrule(lr){8-8}
Dataset & OOD type & Model & AD & PM & GV & kNN & kNN Abs20 \\
\midrule
\multirow{7}{*}{Norman}
 & \multirow{7}{*}{Compositional}
 & GEARS              & \textbf{0.982} & 0.978 & 0.903 & 0.290 & $+5.8\%$ \\
 & & GEARS+scGPT       & 0.981 & 0.982 & 0.909 & $-0.045$ & $+1.7\%$ \\
 & & MLP               & 0.660 & 0.558 & 0.410 & 0.426 & $+3.0\%$ \\
 & & Ridge             & 0.528 & 0.491 & 0.423 & 0.345 & $+1.7\%$\\
 & & Transformer       & 0.627 & 0.384 & 0.336 & $-0.217$ & $-3.7\%$ \\
 & & CPA               & 0.585 & 0.398 & 0.316 & 0.323 & $+4.0\%$ \\
 & & scGPT             & 0.598 & 0.258 & 0.177 & 0.070 & $+0.0\%$ \\
\midrule
\multirow{7}{*}{Joung--Zhang}
 & \multirow{7}{*}{Compositional}
 & Ridge              & \textbf{0.939} & 0.812 & 0.519 & $-0.468$ & $-11.6\%$ \\
 & & MLP               & 0.796 & 0.388 & 0.441 & 0.137 & $-3.2\%$ \\
 & & Transformer       & 0.783 & 0.041 & $-0.038$ & 0.294 & $-3.0\%$ \\
 & & CPA               & 0.767 & 0.268 & 0.091 & 0.141 & $+5.0\%$ \\
 & & scGPT             & 0.778 & 0.021 & 0.179 & 0.306 & $+5.7\%$ \\
 & & GEARS+scGPT       & 0.870 & 0.788 & 0.026 & $-0.041$ & $-1.3\%$ \\
 & & GEARS             & 0.830 & 0.729 & $-0.039$ & 0.094 & $-0.6\%$ \\
\midrule
\multirow{5}{*}{Adamson}
 & \multirow{5}{*}{Single-gene}
 & GEARS              & --- & 0.436 & 0.362 & $-0.275$ & $-3.0\%$ \\
 & & MLP               & --- & 0.926 & 0.903 & 0.415 & $+12.1\%$ \\
 & & Transformer       & --- & $-0.047$ & $-0.085$ & $-0.082$ & $-0.6\%$ \\
 & & CPA               & --- & 0.879 & 0.860 & 0.620 & $+12.7\%$ \\
 & & scGPT             & --- & $-0.090$ & $-0.022$ & 0.181 & $+6.4\%$ \\
\bottomrule
\end{tabular}%
}
\end{table*}

\clearpage
\subsection{Reliability-Aware Fallback}
\label{app:fallback}

Given that reliability signals can rank unreliable predictions, we
finally test whether they have decision-level utility on unseen
combinations. We evaluate a conservative fallback analysis in which
high-risk model predictions are replaced by the additive reference.
Table~\ref{tab:fallback-routing} reports cross-fitted routing results,
with thresholds selected on calibration groups and evaluated on
disjoint held-out groups. GEARS and GEARS+scGPT never outperform the
additive baseline on either dataset, and routing substantially reduces
their errors but generally remains slightly worse than always using the
additive reference. For models with mixed pair-level performance,
routing provides only modest improvements. On Joung--Zhang, kNN
routing reduces Ridge MAE from 0.0187 to 0.0185, prediction magnitude
reduces CPA MAE from 0.0192 to 0.0186, and gene-wise variance reduces
Transformer MAE from 0.0201 to 0.0187, all slightly outperforming the
additive baseline of 0.0190. 
Routed performance remains above the oracle in most cases, indicating
that current signals do not fully capture when fallback is beneficial.
Ridge is the main exception on Norman, where it already outperforms the
additive baseline on most evaluations.

\begin{table*}[h]
\centering
\caption{Cross-fitted conservative fallback analysis. Each column
is a model and each row is a fallback policy. For high-risk predictions,
the model output is replaced by the additive reference; this evaluates
the decision utility of each reliability signal rather than assuming
that the additive reference is universally reliable. Thresholds are
selected on calibration groups and evaluated on disjoint held-out
groups. Oracle selects whichever prediction has lower error for each
evaluation. Wins denotes the number of evaluations where the model
outperforms the additive baseline.}
\label{tab:fallback-routing}
\small
\resizebox{\textwidth}{!}{%
\begin{tabular}{lccccccc}
\toprule
\multicolumn{8}{c}{\textbf{Norman} (75 evaluations)} \\
\midrule
Routing policy & Ridge & CPA & MLP & Trans. & scGPT & GEARS+scGPT & GEARS \\
Wins/Additive & 60/75 & 11/75 & 6/75 & 8/75 & 6/75 & 0/75 & 0/75 \\
\midrule
No routing (always model)
& \textbf{0.0118} & 0.0175 & 0.0214 & 0.0215 & 0.0224 & 0.0338 & 0.0336 \\
Route by additive disagreement
& 0.0122 & 0.0143 & 0.0144 & 0.0143 & 0.0144 & 0.0144 & 0.0144 \\
Route by prediction magnitude
& 0.0122 & 0.0144 & 0.0145 & 0.0145 & 0.0150 & 0.0144 & 0.0144 \\
Route by gene-wise variance
& 0.0122 & 0.0145 & 0.0148 & 0.0145 & 0.0146 & 0.0144 & 0.0144 \\
Route by kNN distance
& 0.0121 & 0.0142 & 0.0144 & 0.0145 & 0.0144 & 0.0160 & 0.0156 \\
No routing (always additive)
& 0.0142 & 0.0142 & 0.0142 & 0.0142 & 0.0142 & 0.0142 & 0.0142 \\
Oracle routing
& 0.0115 & 0.0135 & 0.0139 & 0.0138 & 0.0138 & 0.0142 & 0.0142 \\
\midrule
\multicolumn{8}{c}{\textbf{Joung--Zhang} (44 evaluations)} \\
\midrule
Routing policy & Ridge & CPA & MLP & Trans. & scGPT & GEARS+scGPT & GEARS \\
Wins/Additive & 18/44 & 29/44 & 2/44 & 20/44 & 13/44 & 0/44 & 0/44 \\
\midrule
No routing (always model)
& 0.0187 & 0.0192 & 0.0252 & 0.0201 & 0.0209 & 0.0500 & 0.0512 \\
Route by additive disagreement
& 0.0188 & 0.0192 & 0.0195 & 0.0192 & 0.0192 & 0.0207 & 0.0203 \\
Route by prediction magnitude
& 0.0188 & \textbf{0.0186} & 0.0193 & 0.0189 & 0.0191 & 0.0200 & 0.0200 \\
Route by gene-wise variance
& 0.0190 & \textbf{0.0186} & 0.0193 & \textbf{0.0187} & 0.0192 & 0.0207 & 0.0213 \\
Route by kNN distance
& \textbf{0.0185} & 0.0190 & 0.0190 & 0.0190 & 0.0190 & 0.0203 & 0.0215 \\
No routing (always additive)
& 0.0190 & 0.0190 & 0.0190 & 0.0190 & 0.0190 & 0.0190 & 0.0190 \\
Oracle routing
& 0.0181 & 0.0177 & 0.0189 & 0.0181 & 0.0184 & 0.0190 & 0.0190 \\
\bottomrule
\end{tabular}%
}
\end{table*}

%%%%%%%%%%%%%%%%%%%%%%%%%%%%%%%%%%%%%%%%%%%%%%%%%%%%%%%%%%%%%%%%%%%%%%%%%%%%%%
\clearpage
\section{Limitations and Future Directions}
\label{app:limitations}

The present framework has several limitations. First, additive
disagreement requires observed single-gene effects for both constituent
perturbations. It is therefore defined for compositional OOD, where the
components are individually observed, but not for settings in which the
perturbed gene itself is unseen, such as the Adamson evaluation.

Second, AD measures the magnitude of the model-implied interaction
correction rather than its biological correctness. A large correction
may be accurate when it matches the true interaction, and a small
correction may still be wrong. AD should therefore be interpreted as a
risk signal for unsupported extrapolation, not as a direct estimate of
interaction error.

Third, reliability depends on the evaluation target. AD performs
strongly under full-transcriptome MAE on Norman but transfers poorly to
Top20 error. A global transcriptome-wide score may therefore be
insufficient for applications focused on particular genes, pathways, or
cell states. Gene-level and pathway-level disagreement measures are a
natural extension.

Finally, the present study establishes the framework for pairwise
perturbations across distinct compositional settings. Extending it to
higher-order combinations will require richer references than simple
pairwise additivity, potentially using hierarchical, pathway-aware, or
mechanism-informed priors. The same decomposition may also generalize
to other compositional scientific settings, including drug combinations
and multimodal interventions.

\section*{Declaration of LLM Usage}
Large language models were used solely as writing and coding assistants
during the preparation of this work. Specifically, they were used to
polish prose, improve clarity, and assist with routine coding tasks such
as reformatting tables and debugging plotting and evaluation scripts.

\section*{Ethics Statement}
This work studies the reliability of perturbation-response predictors
using publicly available, de-identified Perturb-seq datasets. It involves
no human subjects, no personally identifiable information, and no data
collection. The study is diagnostic in nature: by identifying when
predictions rely on unsupported interaction extrapolation, it aims to
reduce wasted experimental resources and misleading biological
interpretations in downstream screening pipelines. We are not aware of
foreseeable harmful applications arising directly from this work.

% \section*{Reproducibility Statement}
% We provide anonymized code at \url{https://anonymous.4open.science/r/OOD-D6EE/}. Datasets, preprocessing, data splits (seeds 42--44), and per-model hyperparameters are documented in Appendix~B (Tables~4 and~5), and the additive-reference construction and reliability-signal definitions are given in Section~2 and Appendix~B. All datasets used are publicly available and cited to their original publications. All experiments were run on an internal cluster of 10 NVIDIA A100 GPUs.

% \clearpage
% \input{checklist}

\end{document}